\documentclass[10pt,twocolumn,letterpaper]{article}

\usepackage{wacv}              

\usepackage{graphicx}
\usepackage{amsmath}
\usepackage{amssymb}
\usepackage{booktabs}

\usepackage{multirow}
\usepackage{graphicx}
\usepackage[table]{xcolor}
\usepackage{subcaption}
\usepackage{float}
\usepackage{enumitem}

\usepackage[pagebackref,breaklinks,colorlinks]{hyperref}
\usepackage{hhline}
\usepackage[capitalize]{cleveref}
\crefname{section}{Sec.}{Secs.}
\Crefname{section}{Section}{Sections}
\crefname{table}{Tab.}{Tabs.}
\Crefname{table}{Table}{Tables}
\usepackage[accsupp]{axessibility}  

\def\wacvPaperID{441} 
\def\confName{WACV}
\def\confYear{2024}

\def\mypar#1{\vspace{1mm}{\noindent\bf #1.}\hspace{1mm}}

\newcommand{\dino}{DINOv2\xspace}

\newcommand{\ours}{Guided Distillation\xspace}

\begin{document}

\title{\ours  for Semi-Supervised Instance Segmentation}

\author{Tariq Berrada$^{1,2}$
\and
Camille Couprie$^{1}$
\and
Karteek Alahari$^{2}$
\and
Jakob Verbeek$^{1}$
\and
\\
 {\normalsize $^1$ FAIR, Meta \quad\quad $^2$ Univ.\ Grenoble Alpes, Inria, CNRS, Grenoble INP, LJK}
}

\maketitle

\begin{abstract}    
Although instance segmentation methods have improved considerably, the dominant paradigm is to rely on fully-annotated training images, which are tedious to obtain. 
To alleviate this reliance, and boost results, semi-supervised approaches leverage unlabeled data as an additional training signal that limits overfitting to the labeled samples. 
In this context, we present novel design choices to significantly improve teacher-student distillation models.
In particular, we 
(i) improve the distillation approach by introducing a novel ``guided burn-in'' stage, and 
(ii) evaluate different instance segmentation architectures, as well as backbone networks and pre-training strategies.
Contrary to previous work which uses only supervised data for the burn-in period of the student model, we also use guidance of the teacher model to exploit unlabeled data in the burn-in period. 
Our improved distillation approach leads to substantial improvements over previous state-of-the-art results.
For example, on the Cityscapes dataset we improve mask-AP from 23.7 to 33.9 when using labels for 10\% of images, and 
on the COCO dataset we improve mask-AP from 18.3 to 34.1 when using labels for only 1\% of the training data.
\end{abstract}
\begin{figure}
    \centering
    \includegraphics[width=.83\linewidth]{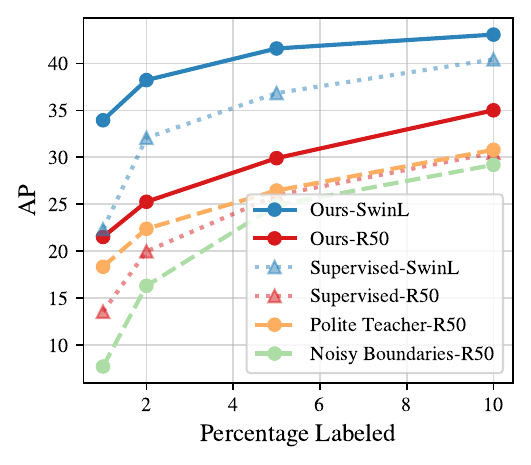}
    \caption{
    Compared with the state-of-the-art Polite Teacher method~\cite{filipiak22arxiv}, we achieve +15.7 mask-AP when using 1\% of labels, for an AP of 34.0, which is more than what Polite Teacher achieved using 10 times more labels (30.8) on the COCO dataset.  
    } 
    \label{fig:ap_comp_coco}
\end{figure}

\begin{figure*}
    \centering
    \includegraphics[width=\linewidth]{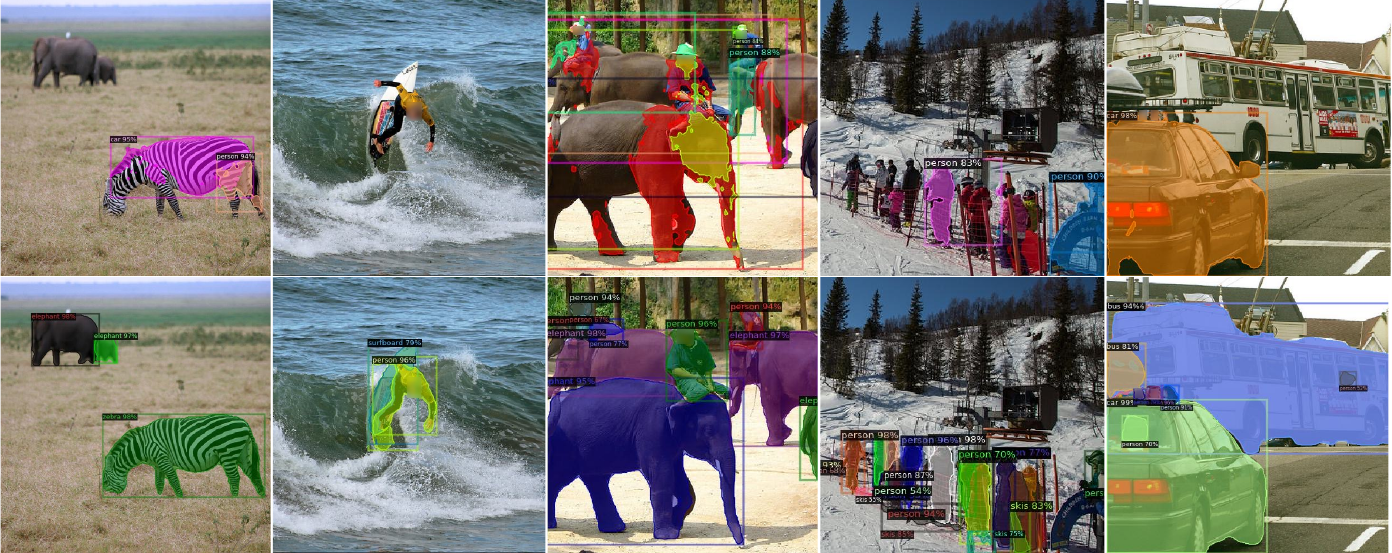}
    \caption{Results on the COCO dataset when using from 5\% labeled data, with supervised training (top) and our semi-supervised approach (bottom). Our approach leads to more detected objects that are segmented with higher accuracy. 
    Results best viewed by zooming in.}
    \label{fig:overview}
\end{figure*}

\section{Introduction}
\label{sec:intro}

The goal of instance segmentation is to extract a segmentation mask and a class label for all relevant object instances in an image.
This generalizes both object detection, for which each instance is localized only with a bounding box, as well as semantic segmentation, in which different instances of the same class can be lumped together in the same mask.
Although recent segmentation models have brought considerable improvements in performance to this task, see e.g.~\cite{he17iccv,mask2former}, these methods require large amounts of images with pixel-level annotations in order to perform well. 
Consequently, the difficulty of manually producing precise instance segmentation maps for training forms a barrier for
applying  such models to settings where relevant labeled data sets are not readily available. 

Semi-supervised learning can be used in label-scarce situations by making use of unlabeled images to boost performance: for the unlabeled images  the  labeling process is avoided, while they still provide useful problem specific input data, see e.g.~\cite{li21cvpr,kipf17iclr,springenberg16iclr,engelen20ml}. 
For semi-supervised instance segmentation, the ``Noisy Boundaries'' method~\cite{wang22cvpr} builds upon the detection-based Mask-RCNN model~\cite{he17iccv}, and leverages pixel-level pseudo labels on unlabeled images, while also making the training of the mask prediction head more robust to noise.
It achieves comparable performance to Mask-RCNN while utilizing only 30\% of labeled images. 
While providing an important gain in label efficiency, this approach relies on a fixed Mask-RCNN teacher model, which is trained in supervised mode from the available labeled images, which limits its performance in settings with few labeled images.

The Polite Teacher approach~\cite{filipiak22arxiv}  is currently state-of-the-art for semi-supervised instance segmentation. 
It performs distillation on a CenterMask2~\cite{lee2020centermask} model with a joint-training framework, where the teacher is updated as an exponential moving average (EMA) of the student model.  
Low confidence thresholding on bounding-box and mask scores is used to filter out noisy pseudo labels. 
In our work, we also adopt a student-teacher training with EMA, with several important differences \wrt the Polite Teacher approach: 
we introduce a  novel  burn-in step in the distillation process, and adopt an improved Mask2Former segmentation  architecture~\cite{mask2former}  in combination with Swin~\cite{liu21iccvswin}  and \dino~\cite{oquab2023dinov2} feature backbones.    

To summarize, our contributions are the following :
\begin{itemize}[noitemsep,topsep=-\parskip]
\itemsep0em 
    \item We develop a  novel teacher-student distillation strategy to more effectively leverage unlabeled images.
    \item We leverage vision transformer architectures for the first time in the context of semi-supervised instance segmentation. 
    \item We show largely improved performance in terms of mask-AP for both Cityscapes and COCO datasets across a wide range of labeled data budgets.
\end{itemize}
Our  implementation can be found on the \href{https://github.com/facebookresearch/GuidedDistillation}{project webpage~\footnote{https://github.com/facebookresearch/GuidedDistillation}}.
\section{Related work}
\label{sec:related}

In this section, we review instance segmentation models and their applications in the semi-supervised regime.

\mypar{Segmentation models}
Instance segmentation aims at identifying the semantic class of every pixel in the image, as well as associating each pixel with a specific instance of the object class. This  contrasts with semantic segmentation which only aims at solving the first part. 
Instance segmentation is more challenging because it requires 
identifying each individual object instance, 
associating parts with the correct objects, and 
accurately determining the boundaries of individual objects. 
These difficulties compound with other challenges, such as 
variability in scale, color, lighting, occlusions, \etc
Until recently, the state-of-the-art was held by detection based approaches such as Mask-RCNN~\cite{he17iccv} and follow-ups such as ViTDet~\cite{li22arxiv}, with a component to detect object bounding-boxes and one for  binary segmentation  to isolate the object from  background pixels.  
Further progress in transformers \cite{maskformer, mask2former} allowed instance segmentation approaches to avoid a separate detection step.
Additional gains have been brought by leveraging text-level supervision using CLIP 
 and very large models~\cite{fang2023eva}. 
To our knowledge, our work is the first one to explore the application of ViTs in the context of semi-supervised instance segmentation.
Recently, MaskFormer~\cite{maskformer} observed that mask classification is sufficiently general to solve both semantic and instance-level segmentation tasks. Their approach converts per-pixel classification into a mask classification model using a transformer-based set prediction mechanism. 
Mask2Former~\cite{mask2former} provides further  improvements and presents a universal segmentation model using the same mask classification mechanism, an efficient masked attention module, multi-scale feature pooling and other improved techniques for fast and efficient training.

\mypar{Semi-supervised instance segmentation}
Previous work has considered the use of pseudo labeling to leverage unlabeled data for semi-supervised instance segmentation~\cite{filipiak22arxiv, wang22cvpr}.
A ``teacher'' instance segmentation model is used to annotate unlabeled images, which are then used by the ``student'' model in addition to labeled images for which ground-truth instance segmentations are available.  
A key factor for the success of such methods is the discrepancy between the teacher and student models,  which ensures that the teacher provides a stable and strong training signal to the student by means of the pseudo labels.
The first condition ---stability--- can be achieved by using a fixed teacher, or using a teacher with the  same architecture as the student and updating it as an exponential moving average (EMA) from the student~\cite{filipiak22arxiv,rolemodels,wang22cvpr}. 
The second condition ---strong training signal--- can be met by using different data augmentations of the same image for the teacher and student model: using strong augmentations for the student and weaker augmentations for the teacher ensures high quality for the pseudo labels of the teacher and a useful training signal for the student being confronted with  hard prediction tasks on strongly augmented training data.

Among the few existing semi-supervised instance segmentation approaches, 
Noisy Boundaries~\cite{wang22cvpr} performs distillation with a fixed Mask-RCNN teacher model trained from labeled data only. The student network is learning a noise-tolerant mask head for boundary prediction by leveraging low-resolution features. 
The Polite Teacher~\cite{filipiak22arxiv} approach is closer to our work and also uses EMA teacher updates. 
In our work we develop an improved burn-in stage, before starting the main semi-supervised training loop, in which we  pre-train the student model from both labeled data as well as unlabeled data using pseudo labels from a fixed teacher model. In contrast, Polite Teacher pre-trains the student from supervised data only. 

\mypar{Knowledge distillation} Our work employs a specific form of knowledge distillation (KD), where knowledge is transferred from the teacher model to the student by providing pseudo-labels for unlabeled samples in an online fashion, while at the same time updating the student progressively  using EMA. Other approaches have explored teacher-free distillation approaches~\cite{10.1007/tf-fd}, efficient N-to-1 matching of feature representations~\cite{liu2023norm}, bridging the gap between online and offline KD methods with a lightweight shadow teacher module~\cite{li2022shake}, and teacher-dependent student architecture search \cite{dong2023diswot} for optimal knowledge flow. 
Such approaches are complementary to our work.

\section{Method}
\label{sec:meth}

\begin{figure}
    \centering
    \includegraphics[width=\linewidth]{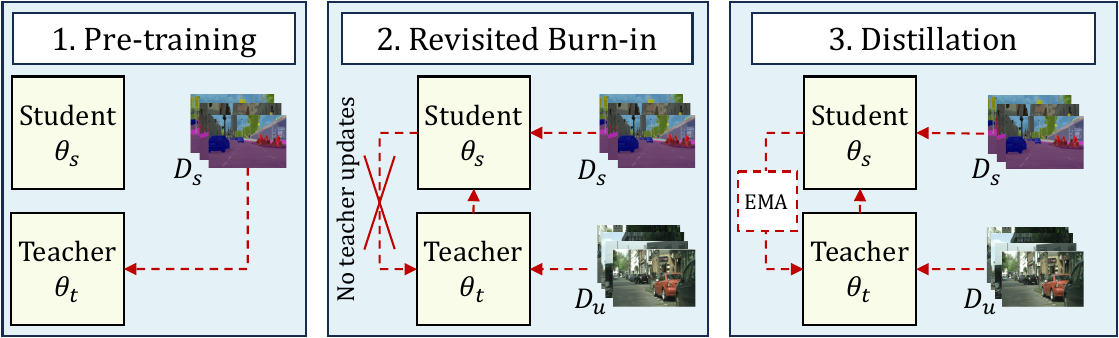}
    \caption{Our semi-supervised training methodology consists of three steps.
    (i) The teacher network is trained on labeled data only. 
    (ii) The student is initialized from scratch, and  trained on both labeled data and unlabeled data using pseudo labels generated by  the pre-trained teacher. 
    The teacher remains fixed during this step. 
    (iii) Copy the checkpoint of the student to the teacher, and then continue training on both labeled and unlabeled data, while using EMA updates for the teacher network from the student's weights.
    }
    \label{fig:diagram_steps}
    \vspace{-2mm}
\end{figure}

\begin{figure*}
    \centering
    \includegraphics[width=.95\linewidth]{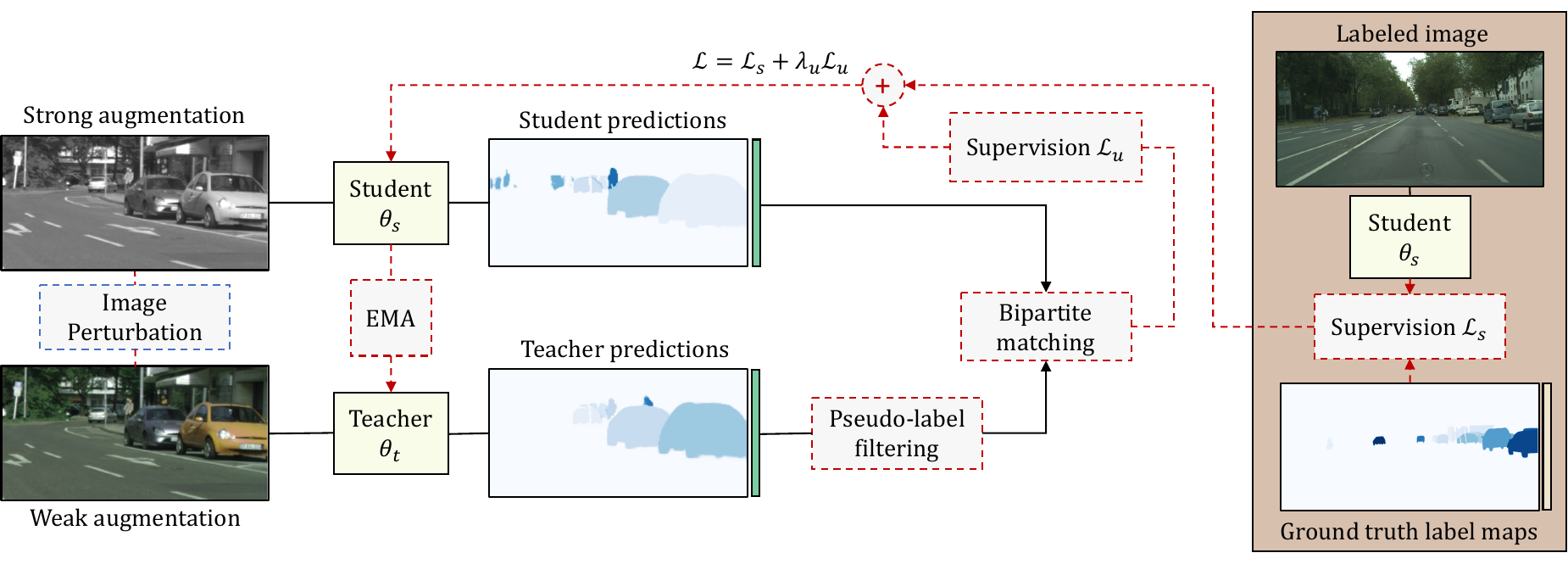}
    \caption{Workflow of our distillation stage. 
    The student and teacher receive weakly and strongly augmented versions of the unlabeled image, respectively. 
    The student predictions are filtered and one-hot encoded to obtain the pseudo-labels that serve as supervision for the student model. 
    We use the same bipartite instance matching method as Mask2Former~\cite{mask2former}. 
    The teacher is then updated using EMA.
    }
    \label{fig:enter-label}
    \vspace{-2mm}
\end{figure*}

Let $D_s = \{(x_1,y_1), ..., (x_N, y_N)\}$ be a supervised dataset and $D_u = \{x'_1, ..., x'_M\}$ an unlabeled set of images, where $x$ denotes images, $y$ their segmentations, and $M \gg N.$ The goal is to make use of $D_u$ in order to improve the performance of the model using the available labels.

Below, in \cref{archi} we present  model architectures, our improved distillation approach, and pseudo labeling strategy.
In \cref{optim} we detail the  losses and EMA teacher updates to train our models.

\subsection{Architectures, distillation and pseudo-labeling}
\label{archi}

\mypar{Model architecture} 
We follow the  Mask2Former architecture~\cite{mask2former} which achieves strong results across different segmentation tasks (semantic, instance, panoptic). 
For our study, we focus on the instance segmentation task. 
The model consists of a backbone for image-level feature extraction, a pixel decoder to gradually upsample image features to compute per-pixel embeddings and a transformer decoder that predicts $N$ mask and class embeddings which generate $N$,  possibly overlapping, binary masks via a dot product between the image-level features and the class embeddings.

\mypar{Teacher-student distillation}
Our training pipeline can be divided into three steps as illustrated in \cref{fig:diagram_steps} :
\begin{enumerate}[noitemsep,topsep=-\parskip]
    \itemsep0em 
    \item {\it Teacher pre-training}: the teacher model, parameterized by $\theta_t$, is trained on annotated data only.
    \item {\it Improved burn-in}: The student model, parameterized by $\theta_s$, is initialized randomly, then trained on both labeled data and unlabeled data using pseudo labels provided by the pre-trained teacher model. The teacher model remains fixed during this stage.
    \item {\it Distillation stage}: In this stage, we  copy the student weights to  the teacher, and then  train the student on both labeled and unlabeled data as before. The teacher  is updated using an EMA of the student's weights.
\end{enumerate}

\mypar{Pseudo-label extraction}
Mask2Former by default predicts a constant number of $K\!=\!100$ masks, which we need to filter in order to obtain pseudo ground-truths for the unlabeled images. To do this, we follow a simple thresholding scheme that takes into account the predicted class probability and the predicted mask size. A predicted mask is selected as a pseudo label if (i) the maximum class probability is above the class threshold $p_c \geq \alpha_C$, and (ii) the size of the predicted mask is above the size threshold: $\sum_{p=0}^{H \times W} \sigma (\hat{y}_{(p)}) \geq \alpha_S$, with $\sigma$ the sigmoid activation of the binary mask prediction. 

\subsection{Optimisation}
\label{optim}
 
In this section we  introduce the loss functions used to train the models, and  present  our guided distillation approach in detail. 
An illustration of the distillation training stage is provided in \cref{fig:enter-label}. 

\mypar{Loss}
Following Mask2Former~\cite{mask2former}, we match predictions from the student  and pseudo ground-truth instances from the teacher by constructing a cost matrix for bipartite matching similarly to DETR~\cite{carion20eccv}. Once the matching is obtained, we use a weighted sum of cross entropy and Dice loss for the masks and cross-entropy loss for the class predictions. In order to minimize memory requirements, we follow PointRend's approach and only sample a limited number of points from the high resolution masks using importance sampling~\cite{pointrend}. Furthermore, this loss is computed for every auxiliary output of the model as in \cite{mask2former}.

Let $x_i$ be an image and $y_i=\{(y^k_i, c^k_i)\}_{k=1,\dots,n}$ its associated ground-truth instances defined by the $n$ binary masks $y_i^k$ and class indices $c_i^k$.
The model predicts $K$ candidate instances $(\hat{y}^k_i, \hat{c}^k_i)_{1 \leq k \leq K}= S_m \circ G(x_i; \theta)$, where $G$ is the backbone and $S_m$ the segmentation head. 
Using bipartite matching, we obtain a matching between the ground truth and predictions $(k, \tau_k)_{1 \leq k \leq n}$, which is used to calculate the image-level loss for both masks and predicted classes:
\begin{eqnarray}
    \mathcal{L}_s^i 
     &\!=\!& \frac{1}{n}  \sum_{k=1}^n l_{CE} (\hat{y}^{\tau_k}_i, y^k_i) + \lambda_D l_D(\hat{y}^{\tau_k}_i, y^k_i) \nonumber\\
     & &  + \lambda_C l_C (\hat{c}_i^{\tau_k}, c_i^k),  \label{Lsi}\\
    \mathcal{L}_s &\!=\!& \frac{1}{|B|} \sum_{i \in B} \mathcal{L}_s^i, \label{Ls}
\end{eqnarray}
where $l_{CE}$ is the cross-entropy loss  for the class predictions, $l_C$ binary cross-entropy and $l_D$ the dice loss function, $\lambda_C$ and $\lambda_D$ are scaling parameters, and $B$ is the  training batch.

For the unsupervised loss, we follow the same procedure as the supervised setting. 
To obtain the pseudo labels, we filter the predictions of the teacher as described previously, and then define the pseudo labels by one-hot encoding both the mask and class predictions. 
Let $x'_j$ be an unlabeled image and $y'_j=(\hat{y}'^k_j, \hat{c}'^k_j)_{1 \leq k \leq K}= S_m \circ G(x'_j; \theta_t)$ the predictions of the teacher, then  pseudo-labels are defined as:
\begin{equation}
    \begin{cases}
        c'^k_j & = \text{argmax}_{c \in C} c^k_j,\\
        y'^k_j & = \sigma ( \hat{y}'^k_j) \geq 0.5.
    \end{cases}
\end{equation}
Given the pseudo labels, we compute the unsupervised loss term $\mathcal{L}_u$ analogous to how we compute the supervised loss $\mathcal{L}_s$ in equations~(\ref{Lsi}) and~(\ref{Ls}) above.
Finally, we combine the supervised and unsupervised loss in a weighted sum as:
\begin{equation}
    \mathcal{L} = \mathcal{L}_s + \lambda_u \mathcal{L}_u.
\end{equation}

\mypar{Revisited burn-in stage}
To ensure  that the quality of the pseudo labels is high enough, existing distillation-based 
semi-supervised methods  adopt a burn-in stage before utilizing the teacher in the training pipeline, \eg as in  Polite Teacher~\cite{filipiak22arxiv}. 
During this stage, the student is trained on labeled data only before copying its weights to the teacher.
However, this is not ideal since the student model will be more likely to converge towards a local minimum induced by its initial training where only the limited labeled data was taken into account. 
Alternatively, we show that it is better to make use of the teacher model from the beginning of training. 
To do this, we first train the teacher model on labeled data only and then use it to provide pseudo labels for training the student model on both labeled and unlabeled data.
During this initial training stage, which we refer as  our revisited burn-in stage, the teacher model stays constant and is therefore not influenced by the student predictions. 
This enables the student model to learn from more data, in particular in cases where very few labeled images are available in combination with many unlabeled images, making it less prone to over-fitting on the limited labeled training data. 
One important hyperparameter is the length of the burn-in stage, which we choose so that the EMA updates of the teacher start when the student's performance becomes comparable or superior to that of the pre-trained teacher.

We recap in \cref{tab:burnin} the different training strategies from the literature and highlight the specificity of our approach.

\begin{table}
    \centering
    \resizebox{\linewidth}{!}{
 \begin{tabular}{llccc}
    \toprule
&     & Polite  & \ours  & Noisy \\
&     & Teacher &  (Ours) & Boundaries\\
    \midrule
     Init 0  & &Random & Pre-train $T$ & Pre-train $T$ \\
    \midrule
    \multirow{2}{*}{Burn-in} & Student &  on $D_s$   &  on $D_s\!+\!D_u$& \\
& Teacher    &   ---  &  Fixed &  \\
     \midrule
     Init 1 & & {$T \leftarrow S$} & {$T \leftarrow S$} & ---\\
     \midrule
     \multirow{2}{*}{Training} & Student &   on $D_s\!+\!D_u$ &  
     on $D_s\!+\!D_u$ & on $D_s\!+\!D_u$ \\
    &  Teacher &   EMA &  EMA &  Fixed \\
    \bottomrule
  \end{tabular}}
    \caption{Comparing our proposed distillation approach to the ones previously used in Polite Teacher~\cite{filipiak22arxiv}  and Noisy Boundaries~\cite{wang22cvpr}. 
    }
    \label{tab:burnin}
\end{table}

\mypar{Exponential moving average (EMA)}
During the second stage of training, the teacher model is updated as an EMA of  the student's weights. This approach has been proven to stabilize training by providing more regular pseudo labels during the student's training~\cite{rolemodels}.
Let $\theta_s^n$ and $\theta_t^n$ be the student and teacher's weights respectively at iteration $n$. The update rule for the teacher  can be written as:
\begin{equation}
    \theta_t^{n+1} = \alpha \cdot \theta_t^n + (1-\alpha) \cdot \theta_s^{n+1},
\end{equation}
where $\alpha$ is the decay rate which regulates the contribution of the student's weights to the update in every iteration.

\section{Experiments}
\label{sec:exp}
Here we present our experimental setup in \cref{setup}, and present our main results followed by ablations in \cref{results}.

\subsection{Experimental setup}
\label{setup}
\mypar{Datasets and evaluation metric}
We use two datasets in our experiments. 
The {\bf Cityscapes} dataset~\cite{cordts16cvpr} that contains 2,975 training images of size $1024\times 2048$ taken from a car driving in German cities, labeled with 8 semantic instance categories. 
Following~\cite{wang22cvpr}, we train models using 5\%, 10\%, 20\% and 30\% of the available instance annotations, and evaluate using the 500 validation images with public annotations. The different data splits are generated by randomly selecting a random subset from the training images. 
For the semi-supervised results we present with $100\%$ annotated images, we use the additional 20k unlabeled images.
The {\bf COCO}~\footnote{The COCO dataset, and all the experiments described in this paper have only been carried out for the purpose of writing this scientific paper and are not to be used in the context of any product or service.} dataset~\cite{lin14eccv} has instance segmentations for 118,287 training images. 
We adopt the same evaluation setting as \cite{filipiak22arxiv}, using 1\%, 2\%, 5\% and 10\% of labeled images for semi-supervised training and using the remaining training images as unlabeled. 
We use the same supervised/unsupervised splits as earlier work~\cite{filipiak22arxiv,liu2021unbiased}. 
Following previous work~\cite{filipiak22arxiv,he17iccv,wang22cvpr}, we use the standard {\bf mask-AP} metric
to evaluate instance segmentation quality.

\mypar{Implementation}
We implement our approach in the Detectron2 codebase~\cite{wu2019detectron2}. 
By default, experiments are conducted on one machine with eight V100 GPUs with 32 GB RAM each. 
In some cases less GPUs were needed, \eg only four GPUs are needed to train semi-supervised models with a ResNet-50 backbone on Cityscapes. 
For experiments using Swin-L, ViT-B or Vit-L backbones on COCO semi-supervised training we needed 16 GPUs across two machines to keep the same batch size. 
For our experiments with these large backbones, we enabled automatic mixed precision in order to reduce the memory footprint.
In \cref{app-models} we provide more details on the sizes, efficiency, and carbon footprint of our models.

\mypar{Hyper-parameters}
Our hyper-parameters are set as follows. We set the EMA decay rate to $\alpha = 0.9996$, the unsupervised loss weight to $\lambda_u = 2$, the class threshold to $\alpha_C = 0.7$ and the size threshold to $\alpha_S = 5$. 
For the loss weights, we follow~\cite{mask2former} and set $\lambda_D=5$ and $\lambda_C=1$.
For Cityscapes, we train our models for 90k iterations, except for our semi-supervised model with the additional 20k images and 100\% of labeled images used, in which case we found it beneficial to double the training length. 
For COCO, we train our models for $368$k iterations. For all our models, we use AdamW with a learning rate of $10^{-4}$, weight decay of $0.05$ and multiplier of $0.1$ for the backbone updates, a batch size of $16$ and early stopping. 
The number of iterations used for the burn-in period is given in \cref{tab:burn_in_l_city}.
\begin{table}
    \centering
        \resizebox{\linewidth}{!}{
    \small
    \begin{tabular}{lccccccc}
    \toprule
        Labels used  & 1\% & 2\% & 5\% & 10\% & 20\% & 30\% & 100\%\\
        \midrule
         Cityscapes & --- & --- & 15k & 25k & 30k & 35k & 50k\\
         COCO       & 20k & 25k & 30k & 60k & --- & --- & ---\\
        \bottomrule
    \end{tabular}
    }
    \caption{Number of burn-in iterations used in our approach.}
    \label{tab:burn_in_l_city}
\end{table}

\subsection{Main results}
\label{results}

\begin{figure*}
    \includegraphics[width=\linewidth]{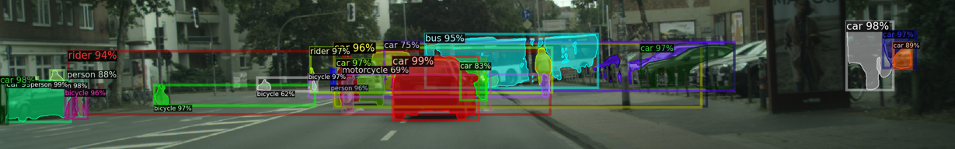}\\
    \includegraphics[width=\linewidth]{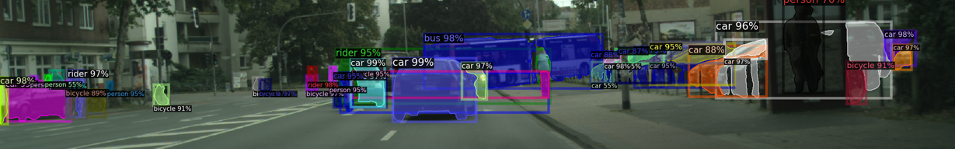}
    \caption{Overview of our results on Cityscapes with 10\% labeled data with R50 backbone. {\bf Top:} Predictions of model trained using $5\%$ of labeled data only. {\bf Bottom :} Predictions of model trained using our semi-supervised method using $5\%$ of labeled data. Without our supervision method, the model tends to merge different instances from the same semantic category (bounding boxes spanning multiple objects), the model also misidentifies certain objects  by predicting a the wrong class or only a certain portion of the instance. 
    Our method greatly helps in alleviating all these issues.}
    \label{fig:city_illustr}
\end{figure*}

\begin{table*}
\small
    \begin{subtable}[b]{\columnwidth}
        \centering
        \resizebox{\columnwidth}{!}{%
        \begin{tabular}{lrrrrr}
        \toprule
             Amount of labeled data used & 5\% & 10\% & 20\% & 30\% & 100\% \\
             \midrule
             Mask-RCNN - R50 (IN-1k) \cite{he17iccv}& 11.3 & 16.4 & 22.6 & 26.6 & 33.7\\   
             \midrule
             Mask2Former - R50 (IN-1k)  & 12.1 & 18.8 & 27.4 & 29.6 & 36.6\\
             Mask2Former - ViT-B (\dino)  & 18.6 & 21.9 & 28.1 & 31.1 & 36.5\\
             Mask2Former - ViT-L (\dino)  & {\bf 22.9} & {\bf 27.1} & 31.4 & 33.7 & 38.0\\
             Mask2Former - Swin-L (IN-21k)   & 17.8 & 26.3 & {\bf 33.2} & {\bf 37.1} & {\bf 43.5}\\
             \bottomrule
        \end{tabular}
        }
        \caption{ {\bf Supervised models on Cityscapes}.}
        \label{cityscapes_sup}
    \end{subtable}
    \hfill
    \begin{subtable}[b]{\columnwidth}
        \centering
        \resizebox{\columnwidth}{!}{
        \begin{tabular}{lrrrrr} 
        \toprule
         Amount of labeled data used     & 1\% & 2\% & 5\% & 10\% & 100\% \\
             \midrule
             Mask-RCNN - R50 (IN-1k) \cite{he17iccv} & 3.5 & 9.5 & 17.4 & 21.9 & 37.1\\
              CenterMask2 - R50 (IN-1k) \cite{lee2020centermask} & 10.1 & 13.5 & 18.0 & 22.1 & ---\\
              \midrule
              Mask2Former - R50 (IN-1k)  & 13.5 & 20.0 & 26.0 & 30.5 &  43.5\\
              Mask2Former - ViT-B (\dino)    & 25.9 & 30.0 & 33.6 & 36.3 & 47.0\\
              Mask2Former - ViT-L (\dino)  & {\bf 30.2} & {\bf 33.1} & 36.6 & 39.1 & 47.9\\
              Mask2Former - Swin-L (IN-21k)  & 22.3 & 32.1 & {\bf 36.9} & {\bf 40.4}  & {\bf 50.0}\\
             \bottomrule
        \end{tabular}}
        \caption{{\bf Supervised models on COCO}.
        }
        \label{coco_sup}
    \end{subtable}\\
    \begin{subtable}[b]{\columnwidth}
        \resizebox{\columnwidth}{!}{%
        \small
        \begin{tabular}{lrrrrr}
        \toprule
             Amount of labeled data used & 5\% & 10\% & 20\% & 30\% & 100\% \\
             \midrule
             Noisy Boundaries - R50 (IN-1k) \cite{wang22cvpr} & 21.2 & 23.7 & 30.8 & 33.2 & 34.7\\
             \cmidrule{1-6}
             Ours - R50 (IN-1k) & 23.0 & 30.8 & 33.1 & 35.6 & 39.6\\
             Ours - ViT-B (\dino) & 25.1 & 27.0 & 34.6 & 35.4 & 39.6\\
             Ours - ViT-L (\dino) & {\bf 28.8} & 33.0 & 36.8 & 39.1 & 42.9\\
             Ours - Swin-L (IN-21k) & 25.1 & {\bf 33.9} & {\bf 38.1} & {\bf 39.6} & {\bf 43.8}\\
             \bottomrule
        \end{tabular}
        }
        \caption{ {\bf Semi-supervised models on Cityscapes}.
        }
        \label{cityscapes_semi}
    \end{subtable}
    \hfill
    \begin{subtable}[b]{\columnwidth}
                \resizebox{\columnwidth}{!}{%
        \small
        \begin{tabular}{lrrrr}
        \toprule
             Amount of labeled data used  & 1\% & 2\% & 5\% & 10\% \\
             \midrule
              DD - R50 (IN-1k) \cite{DD} & 3.8 & 11.8 & 20.4 & 24.2 \\
              Noisy Boundaries - R50 (IN-1k) \cite{wang22cvpr} & 7.7 & 16.3 & 24.9 & 29.2 \\
              Polite Teacher - R50 (IN-1k) \cite{filipiak22arxiv} & 18.3 & 22.3 & 26.5 & 30.8 \\
              \midrule
              Ours - R50 (IN-1k)  & 21.5 & 25.3 & 29.9 & 35.0 \\
              Ours - ViT-B (\dino) & 30.5 & 34.3 & 37.0 & 39.2 \\
              Ours - ViT-L (\dino)& {\bf 34.1} & 37.4 & 40.3 & 42.0 \\
              Ours - Swin-L (IN-21k)  & 34.0 & {\bf 38.2} & {\bf 41.6} & {\bf 43.1} \\
             \bottomrule
        \end{tabular}
        }
        \caption{{\bf Semi-supervised models on COCO}.}
        \label{coco_semi}
    \end{subtable}
    
\caption{Comparison of results on COCO and Cityscapes. Results for Mask-RCNN, CenterMask2, Polite Teacher and Noisy Boundaries are taken from \cite{filipiak22arxiv} and \cite{wang22cvpr}. For each backbone, we indicate the pre-training data, \dino is SSL pre-trained on a dataset of 142M images.}
\end{table*}

\mypar{Semi-supervised instance segmentation results}
In \cref{cityscapes_semi} we compare our results on the Cityscapes dataset to these of Noisy Boundaries (NB)~\cite{wang22cvpr}. 
We provide results using different feature backbones, including the ImageNet-1k pre-trained ResNet-50 backbone used in NB. 
We observe large improvements over the latter using all backbones and percentages of labels used, with best results obtained using the larger SSL \dino and the supervised Swin-L backbones.
The \dino backbone leads to best results (28.8 \vs 25.1 for Swin) in the most challenging setting where we used only 5\% of the labeled data, while the Swin backbone improves by 0.5 to 1.3 points AP in the other settings. 
Notably, compared to NB we improve the AP by more than 10 points from 23.7 to 33.9 when using 10\% of labels.

In \cref{coco_semi} we report results for the COCO dataset, where we compare to Noisy Boundaries (NB)~\cite{wang22cvpr} and Polite Teacher (PT)~\cite{filipiak22arxiv} approaches. 
We similarly observe substantial improvements across all backbones and percentages of labeled data used.
For example, improving the 22.3 AP of PT to 38.2 in the case of using 2\% of the labels.
In this case the Swin-L backbone yields (near) optimal results.
Our results are better overall, but in particular for small amounts of labeled data, which makes the performance curve as a function of the amount of labeled data flatter, see also \cref{fig:ap_comp_coco}.
For example, between the 1\% and 10\% label case, NB improves from 7.7 to 29.2 AP (+21.5), and PT improves from 18.3 to 30.8 (+12.5), where in our case with the Swin-L backbone the performance increase is 9.1 AP points (34.0 \vs 43.1).

On COCO, Wang \etal~\cite{wang22cvpr} also report a semi-supervised experiment using 100\% of labeled data, and adding another 123k unlabeled COCO images. 
In this setting they obtain 38.6 AP, which is similar to the 38.2 AP that we attain using only 2\% labeled data, so using roughly 50$\times$ less labeled images and twice less unlabeled images.

\mypar{Comparison to supervised baselines}
In \cref{cityscapes_sup}, \ref{coco_sup} we present results obtained using supervised training only, to assess the improvements brought by our semi-supervised approach \wrt this baseline for different backbone architectures. 
We also compare to the supervised baselines using the Mask-RCNN and CenterMask2 segmentation models that underlie Noisy Boundaries and Polite Teacher, respectively.

Comparing the supervised and semi-supervised results for Cityscapes in \cref{cityscapes_sup}, \ref{cityscapes_semi}, and similarly for COCO in \ref{coco_sup}, \ref{coco_semi}, we observe that our approach shows consistent improvements over the supervised baseline across all experiments.
Using the Swin-L backbone on Cityscapes in the 5\% labeled data case, we improve the AP by 7.3 points from 17.8 to 25.1 AP, while for the \dino backbone we improve it from 22.9 to 28.8 (+5.9 AP).
For COCO at 1\% labeled data, our semi-supservised approach improves the supervised result of 22.3 to 34.0 (+11.7 AP) using Swin-L, and from 30.2 to 34.1 (+3.9 AP) using the \dino backbone.
See \cref{fig:ap_comp_coco} for a graph comparing our semi-supervised results to the supervised baseline as well as previous methods on COCO.
We provide qualitative examples of instance segmentation using the supervised baseline and our approach on the COCO dataset in \cref{fig:overview} and for Cityscapes in \cref{fig:city_illustr}.
In \cref{COCO-few} we provide additional experiments for the COCO dataset using even smaller labeled training datasets.

\subsection{Interpretation and ablation studies} \label{interp_abl}
In the following, unless specified otherwise,  ablations are performed on Cityscapes with 5\% labeled data and a ResNet-50 backbone. Results are reported on the validation set.
Notably, these experiments allow to pinpoint the performance differences \wrt the state-of-the-art Polite Teacher approach (for which there is no public codebase), while using the same segmentation and backbone architecture.

\mypar{Revisited burn-in stage}
In order to quantify the performance gains obtained by our revisited burn-in stage in isolation of other architectural choices, we compare it to the strategies used by Noisy Boundaries~\cite{wang22cvpr} and Polite Teacher~\cite{filipiak22arxiv}. 
See \cref{tab:burnin} for a comparison. 
We also consider a fourth variant, which is identical to our approach but setting the number of burn-in iterations to zero.
Finally, we include the result when using supervised training only as a baseline result.
The results in \cref{tab:burn_in_abl} show that our approach brings about $+3.7$ AP in comparison with the (standard) burn-in strategy used by Polite Teacher, and $+4.1$ AP over keeping the teacher fixed during training as in Noisy Boundaries.
Dropping the burn-in stage all together gives worst results (16.9 AP), which underlines the importance of this burn-in stage in distillation approaches.  

\begin{table}
    \centering
    \small
    \begin{tabular}{lc}
         \toprule
         Burn-in method & mask AP \\
         \midrule
         Supervised baseline & 12.1\\
        \midrule
         No burn-in & 16.9\\
         Fixed Teacher \cite{wang22cvpr} & 18.9\\
         Standard\cite{filipiak22arxiv} & 19.3\\
         Ours & \bf{23.0}\\
         \bottomrule
    \end{tabular}
    \caption{Influence of burn-in stage on segmentation performance.}
    \label{tab:burn_in_abl}
\end{table}

\begin{figure}[t]
\centering
\includegraphics[width=.83\linewidth]{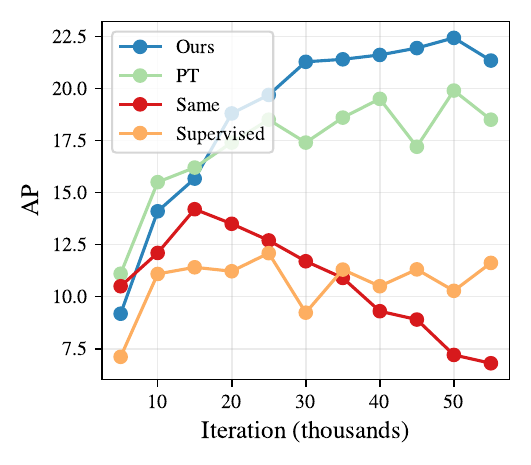}
\caption{Effect of  data augmentation on Cityscapes with $5\%$ labeled data. 
{\bf Ours}: Our default setup which includes color jitter, random grayscale and random blur.
{\bf PT}: Uses {\it Ours} plus cutout as in \cite{filipiak22arxiv}.
{\bf Same}: Teacher and student receive weakly augmented images.
{\bf Supervised}: Model trained with labeled samples only.
}
\label{fig:data_aug}
\vspace{-2mm}
\end{figure}

\mypar{Data augmentations} 
One important component of teacher-student distillation is the asymmetry between the two models that ensures that the teacher yields useful pseudo-labels for the student~\cite{rolemodels}. To fulfill this condition, the teacher and student receive differently augmented versions of the unlabeled image. For the student model, the augmentations consist of random sized crops and horizontal flipping. For the strong augmentations, we add textural-based augmentations, including color jitter, random gray-scale and blur. We show in \cref{fig:data_aug} that this works better than the augmentations used in Polite Teacher~\cite{filipiak22arxiv} which include random cutout, since small instances are prone to be masked which results in inconsistencies in the pseudo-labels provided to the student. 
Without augmentations, the student model quickly diverges while using cutout underperforms compared to textural based augmentations only.

\mypar{Weighting of unsupervised loss}
We compare models trained with different values for the weight $\lambda_u$ for the unlabeled loss term on Cityscapes training with 10\% of labels.
The results in \cref{tab:unlabel_weight} show that even with a low $\lambda_u$, the results are superior to the supervised baseline ($\lambda_u=0$), but the model benefits from putting more emphasis on the semi-supervised loss with larger $\lambda_u$ values. 
However, if this emphasis is too strong then the model tends to overemphasize the teacher's signal to the detriment of the supervised signal, we find it optimal to use $\lambda_u = 2$.

\begin{table}
    \centering
    \small
    \begin{tabular}{cccccc}
    \toprule
         $\lambda_u$ & 0 & 0.1 & 0.5 & 2 & 5 \\
         \midrule
         AP & 18.8 & 23.6 & 24.4 & 30.8 & 25.5 \\
         \bottomrule
    \end{tabular}
    \caption{Influence on unlabeled loss weight on validation performance, on Cityscapes dataset.
    Mask2Former, RN-50, 10\% labels.
    }
    \label{tab:unlabel_weight}
\end{table}

\mypar{Backbone pre-training} 
Most segmentation models rely on pre-trained backbones to extract features on which the segmentation head operates. 
We explore the effect of different datasets to pre-train the feature backbone. 
To reduce computational costs, we perform this exploration using a Swin-B architecture rather than the heavier \mbox{Swin-L} architecture we used in our main experiments. 
In \cref{tab:backbone_pretraining} we report results obtained using ImageNet-1k and ImageNet-21k~\cite{imagenet1k} pre-training for supervised and semi-supervised instance segmentation. 
We find that \mbox{ImageNet-21k} pre-training consistently outperforms pre-training on the smaller ImageNet-1k. 
The performance gap is higher in the low annotation regime at $2\%$ labeled where the improvement is of 4.0 and 4.1 AP points in the (semi)supervised setups.
This difference reduces to 1.4 points in the supervised setting with $100\%$ of labels, indicating a higher importance of the backbone pre-training in regimes with sparser labeling.

\begin{table}
    \centering
    \small
    \begin{tabular}{llccc}
    \toprule
        Pre-training  & Training & $2\%$ & $100\%$\\
        \midrule
        ImageNet-1k & Supervised & 26.3 & 46.8 \\
        Imagenet-21k & Supervised & 30.3 & 48.2\\
        \midrule
        ImageNet-1k & Semi-supervised & 32.4 & --- \\
        Imagenet-21k & Semi-supervised & 36.5 & ---\\
        \bottomrule
    \end{tabular}
    \caption{Backbone pre-training effect on COCO using a Swin-B. 
    }
    \label{tab:backbone_pretraining}
    \vspace{-2mm}
\end{table}

\mypar{SSL backbones} 
We quantify the performance of self-supervised models trained on large scale datasets for our task. More specifically, we adapt ViT~\cite{dosovitskiy21iclr} models trained using \dino~\cite{oquab2023dinov2} and compare the performance obtained with a similar model trained using the Deit~\cite{touvron2021going} method on ImageNet-1K. 
We used similar ViT models, however the \dino model has a patch token size of 14 while the standard ViT has a patch token size of 16.
To adopt ViT models as backbones for our model, we adapt the feature representation of these models before the task-specific head. More precisely, we rearrange the patch tokens to their relative spatial coordinates then use bi-linear interpolation followed by a $1\!\times\!1$ convolution to produce a multiscale feature pyramid.
Results are reported in \cref{coco_sup} and \cref{coco_semi}. 
We see that models trained with \dino backbones show high performance and are comparable with Swin backbones. Such models are also more robust at very sparse regimes, yielding state-of-the-art results on COCO with 1\% of labels and also on Cityscapes at 5\%. However the model does not generalize as well to larger amounts labeled samples, and the gap in performance between Swin-L and \dino-L gets bigger the more labeled samples are available. 
This could be explained by the relatively simple adaptation of the ViT backbone to the segmentation head and the absence of a multi-scale feature pyramid. 

\begin{table}
    \centering
    \small
    \begin{tabular}{llcc}
    \toprule
         Backbone & Model & 2\% & 100\% \\
         \hline
         \multirow{3}{*}{Finetuned} & Deit ViT-B & 20.3 & ---\\
          & \dino ViT-B & 30.0 & 46.9\\
          & \dino ViT-L & 33.1 & 46.7\\
         \midrule
         \multirow{2}{*}{Fixed} & \dino ViT-B & 27.7 & 41.2\\
          & \dino ViT-L & 29.7 & 41.5\\
         \bottomrule
    \end{tabular}
    \caption{Influence of backbone finetuning on training performance using \dino backbone. COCO supervised training.
    }
    \label{tab:dino_fix}
    \vspace{-2mm}
\end{table}

We also experiment with keeping the \dino backbone fixed, and only training the segmentation head to evaluate the feature generalisation ability of the backbone. Finally, to validate whether the robustness at low annotation regimes is a property of the backbone pre-training or its architecture we report results of a Vit-B model trained on ImageNet-1k following Deit~\cite{touvron2021going}. 
As reported in \cref{tab:dino_fix}, we see that the Vit-B Deit pre-trained model is severely underperforming compared to \dino pre-training, and performs on par with a ResNet-50 backbone despite being much larger. This indicates that the performance gains are more due to the large scale pre-training than the ViT's architecture. On the other hand, fixing the backbone and training the segmentation head results in mask-AP values lower  by around $3-4$ points at $2\%$ and $5$ points at $100\%$ supervision. 
This demonstrates that, while not ideal, training a segmentation head on top of frozen large scale pre-trained models such as in \dino can be beneficial when having few annotated samples, especially if available GPU memory is limited.

\section{Conclusions}
\label{sec:concl}

In this paper we introduced \ours, a new  distillation appraoch for semi-supervised instance segmentation. 
This novel approach, combined with recent advances 
in segmentation methods and vision feature backbones,
outperforms previous state-of-the-art in instance segmentation by a large margin on the Cityscapes and COCO benchmarks. 
We found that the optimal pretrained feature backbone varied as a function of the amount of labeled training data: while Swin-L (IN-21k) was optimal for most experiments, we found the ViT-L (DINOv2) backbones to be more effective in the very low  annotation regime. 

In future work, we would like to test our approach for other prediction tasks such as object detection, to explore large scale SSL pretraining of transformer architectures that are better suited for dense prediction, and assess the robustness of our approach to domain shifts.  

\mypar{Acknowledgements} Karteek Alahari is supported in part by the ANR grant AVENUE (ANR-18-CE23-0011).



{\small
\bibliographystyle{ieee_fullname}
\bibliography{biblio, jjv}
}
\newpage
\begin{figure*}
    \centering
    \includegraphics[width=.9\linewidth]{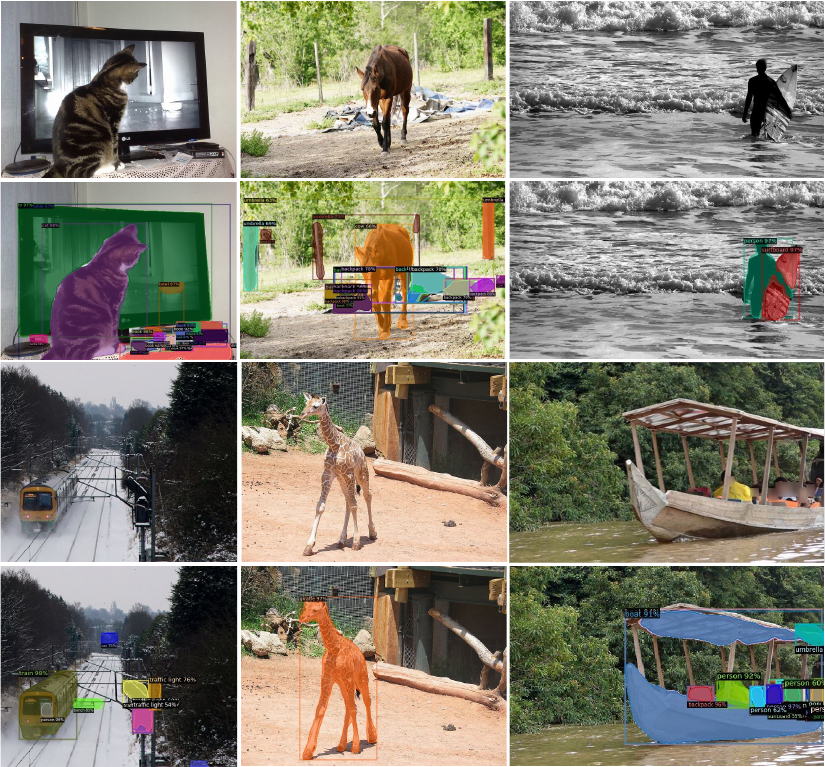}
    \caption{Illustration of predictions obtained with model trained on COCO with only 0.4\% of labels. The model uses a DINOv2 \cite{oquab2023dinov2} backbone and achieves an AP of 31.0, which is superior to what the previous SOTA achieved using 25x more labels  (PT achieves 30.8 AP with 10\% of labels).}
    \label{fig:example_250}
\end{figure*}

\appendix

\section*{Appendix}

\section{COCO experiments with fewer labels}
\label{COCO-few}

We conducted additional experiments on the COCO dataset using fewer labeled training samples compared to earlier work: from 0.4\% down to 0.1\%. 
We report results for a supervised training baseline and our semi-supervised approach in \cref{tab:COCOsmall}.
In this extremely challenging, very sparsely labeled data regime, our approach shows marked improvements over the supervised baseline, \eg, +3.4 points for 0.1\% labeled images using the DINOv2 backbone.
Moreover, we find that in this regime the DINOv2 backbone is more effective than the Swin-L (IN-21k) backbone, \eg, in the 0.4\% labeled data case, corresponding to less than 500 of image annotations: the DINOv2 backbone improves mask-AP by 8 points from 23.0 to 31.0, reaching similar performance as the previous state-of-the-art approach Polite Teacher with 25 times less annotations (30.8 mask AP at 10\% labeled data, see \cref{coco_semi}. 
This is in line with our observation on the 5\% labeled data case for Cityscapes in \cref{cityscapes_semi}, and 1\% labeled data case for COCO in \cref{coco_semi}.
See \cref{fig:example_250} for an illustration of segmentations obtained with this model.

\begin{table}[h]
\centering
\small
\begin{tabular}{lrrr}
\toprule
Amount of labeled data used & 0.1\% & 0.2\% & 0.4\% \\
\midrule
& \multicolumn{3}{c}{Supervised models}\\
\midrule
Mask2Former - Swin-L (IN-21k)      & 5.3  & 10.2 & 15.9  \\
Mask2Former -  ViT-L (DINOv2)	   & \bf 10.2 & \bf 19.4 & \bf 25.7 \\
\midrule
& \multicolumn{3}{c}{Semi-supervised models}\\
\midrule
Ours - Swin-L (IN-21k)      & 5.8  & 16.1 & 23.0 \\
Ours - ViT-L (DINOv2)	    & \bf 13.6 & \bf 24.9 & \bf 31.0 \\
\bottomrule
\end{tabular}
\caption{Evaluation of supervised and semi-supervised models on COCO using extra small labeled training sets.}
\label{tab:COCOsmall}
\end{table}

\section{Details on models and training efficiency}
\label{app-models}

\mypar{Training efficiency}
Our approach can be seen as a two-stage knowledge distillation method where the teacher and student share the same architecture.
Compared to PT \cite{filipiak22arxiv}, we pretrain the teacher network using the available labeled samples only and use its predictions during the student's burn-in stage. 
Compared to NB \cite{wang22cvpr}, we generate pseudo-ground truths in an online manner instead of doing it offline prior to the student's training. 
These changes, although computationally demanding, are justified by the large performance improvements over the previous baselines. 
For example, the student training with ResNet-50 backbone on COCO consumes $89 \%$ more GPU memory and the iterations take approximately twice longer compared to the teacher pre-training.
Peak performance is usually achieved after a few thousand iterations in sparse regimes. 
Therefore, in practice, it only takes a few dozen hours to train the models end-to-end.

Recent works in KD have explored alternative strategies where no separate teacher network is required, see \eg~\cite{10.1007/tf-fd}, with applications to image classification. 
Such alternatives present interesting directions of future work to improve the efficiency of semi-supervised methods.

\mypar{Model characteristics}
We present the different characteristics of the models used. We report both the number of parameters and the FLOP count for each architecture used in our project in Tab.\ \ref{tab:caracteristics} as measured using \verb|count_flops| function in the Detectron2 library.

\begin{table}
    \centering
    
    \begin{tabular}{lrc}
        \toprule
        Backbone &  Parameters (M) & GFLOPS\\
        \midrule
         R50 & 44 & $224.8 \pm 24.6$\\
         Swin-B & 107 & $464.2 \pm 48.7$\\
         Swin-L & 216 & $864.7 \pm 90.2$\\
         ViT-B (DINOv2) & 108 & $944.5 \pm 93.4$\\
         ViT-L (DINOv2) & 326 & $1285.3 \pm 127.1$\\
         ViT-B (Deit) & 108 & $692.8 \pm 98.6$\\
         \bottomrule
    \end{tabular}
    \caption{FLOP and parameter count for the different models used in our project.}
    \label{tab:caracteristics}
\end{table}

\mypar{Comparing training protocols} 
Section \ref{interp_abl} provides ablation for the training protocol, which compares the main changes in the distillation strategy with respect to PT while keeping the overall model architecture constant. Particularly, we :
\begin{enumerate}
    \item Isolate the effect of our revisited burn-in stage.
    \item Isolate the effect of student data augmentations (the teacher augmentations are the same between PT and ours).
\end{enumerate}
In Tab. \ref{tab:burn_in_abl}, we can see that using the standard burn-in stage as in PT  reduces the AP by 3.7 points. In Fig. \ref{fig:data_aug}, we track the mask-AP evolution when using different data augmentations, we can see that our data augmentation yields better performance and more stable convergence that PT which additionally uses random cutout.
Hence, both changes show improved performance with respect to the SOTA protocol while using the same underlying meta architecture, backbone, training epochs etc. This is evidence that our distillation protocol and revisited burn-in stage are both important factors for the improved performance,  beyond the  backbone and meta-architecture choices.

\mypar{Estimation of carbon footprint} 
On COCO, it took 25 hours to train our ViT-L (DINOv2) model using 1\% of annotations, and about 2 days to train a Swin-L (IN-21k) model using 10\% of annotations. 
Trainings are approximately 2.5 times faster on Cityscapes. 
Given the same formula used in  \cite{oquab2023dinov2}, a Thermal Design Power (TDP) of V100-32G equal to 250W, a Power Usage Effectiveness (PUE) of 1.1, a carbon intensity factor of 0.385 kg CO$_2$ per KWh, a time of 2 days $\times$ 24 hours $\times$ 16 GPUs = 768 GPU hours to train our approach with a SWIN-L, it leads to 211 kWh, an equivalent CO$_2$ footprint of 211 $\times$ 0.385 = 81.2 kg.

\end{document}